%% file: main.tex
\documentclass[10pt,twocolumn,letterpaper]{article}

\usepackage[pagenumbers]{cvpr} 

\input{preamble}

\definecolor{cvprblue}{rgb}{0.21,0.49,0.74}
\usepackage[pagebackref,breaklinks,colorlinks,citecolor=cvprblue]{hyperref}
\usepackage{caption}
\usepackage{multirow}
\usepackage{booktabs}
\usepackage{threeparttable}
\def\confName{3DV\xspace}
\def\confYear{2024\xspace}

\title{Towards Loose-Fitting Garment Animation via Generative Model of Deformation Decomposition}
\author{Yifu Liu\\
Alibaba\\
Hangzhou China\\
{\tt\small zhencang.lyf@alibaba-inc.com}
\and
Xiaoxia Li\\
Alibaba\\
Hangzhou China\\
{\tt\small suian.lxx@alibaba-inc.com}
\and
Zhiling Luo\\
Alibaba\\
Hangzhou China\\
{\tt\small godot.lzl@alibaba-inc.com}
\and
Wei Zhou\\
Alibaba\\
Hangzhou China\\
{\tt\small fayi.zw@alibaba-inc.com}
}

\begin{document}
\maketitle
\input{sec/0_abstract}    
\input{sec/1_intro}
\input{sec/2_related}
\input{sec/3_method}
\input{sec/4_expr}
\input{sec/5_conc}
{
    \small
    \bibliographystyle{ieeenat_fullname}
    \bibliography{main}
}

\end{document}

%% file: preamble.tex
\usepackage[dvipsnames]{xcolor}


%% file: sec/0_abstract.tex
\begin{abstract}
Existing data-driven methods for garment animation, usually driven by linear skinning, although effective on tight garments, do not handle loose-fitting garments with complex deformations well.
To address these limitations, we develop a garment generative model based on deformation decomposition to efficiently simulate loose garment deformation without directly using linear skinning.
Specifically, we learn a garment generative space with the proposed generative model, where we decouple the latent representation into unposed deformed garments and dynamic offsets during the decoding stage. 
With explicit garment deformations decomposition, our generative model is able to generate complex pose-driven deformations on canonical garment shapes.
Furthermore, we learn to transfer the body motions and previous state of the garment to the latent space to regenerate dynamic results.
In addition, we introduce a detail enhancement module in an adversarial training setup to learn high-frequency wrinkles. 
We demonstrate our method outperforms state-of-the-art data-driven alternatives through extensive experiments and show qualitative and quantitative analysis of results.
\end{abstract}

%% file: sec/1_intro.tex
\section{Introduction}
\label{sec:intro}

Garments play an essential role in our daily lives.
Nowadays, more and more applications such as shopping, video games, virtual try-on, and virtual reality require realistic garment dynamics, especially for loose-fitting garments.
In the traditional computer graphic (CG) industry, Physical-Based Simulations (PBS) \cite{nealen2006physically} are addressed to model realistic garment deformations, but they are computationally expensive and often require expert knowledge to complete cloth material properties.

\begin{figure}[t]
\vspace{-0.0cm} 
\setlength{\abovecaptionskip}{0.2cm}  
\setlength{\belowcaptionskip}{-0.5cm} 
    \centering
    \includegraphics[width=1.0\linewidth]{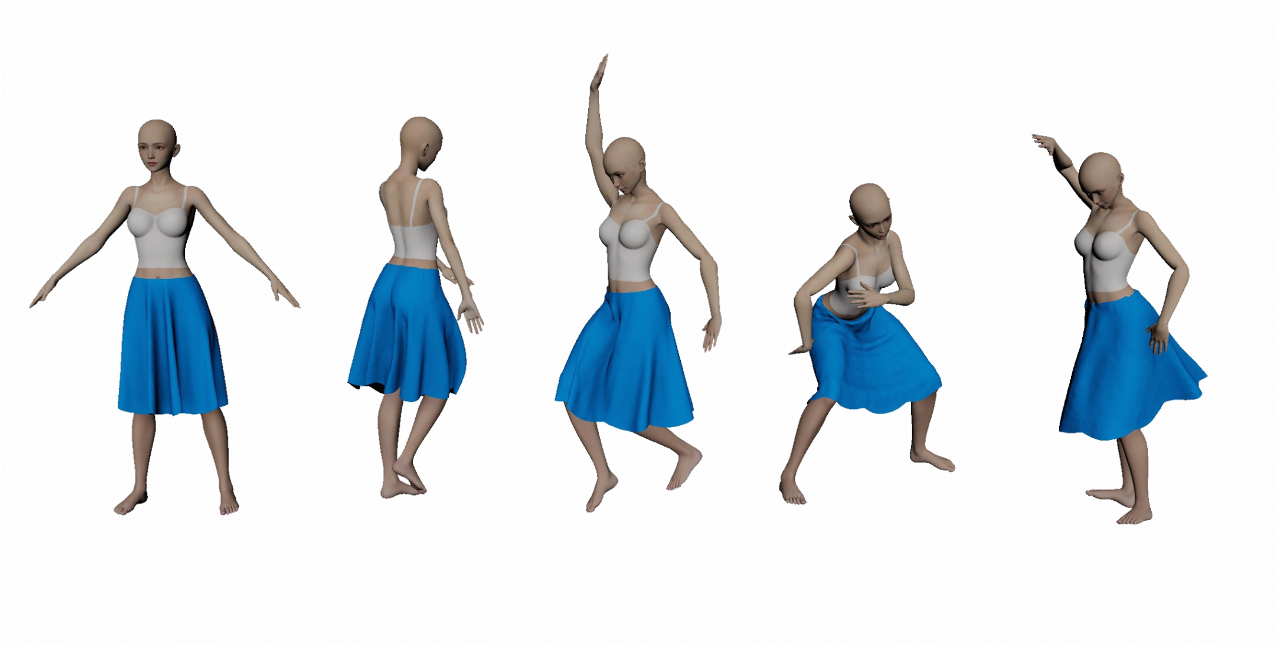}

    \caption{
    We present a novel framework for garment animation.
    Our method predicts dynamic deformed garments based on the learned generative model, and for loose-fitting garments, 
    we can simulate the free-flowing shape of skirts under large-scale motions (from left to right).
    }
    \label{fig1}
\end{figure}

Recently, data-driven or learning-based methods have emerged with promising results to approximate PBS methods.
The core idea is to learn a neural network that infers garment deformations under target motions based on the observation in a large dataset.
However, most of them \cite{santesteban2019virtualtryon,santesteban2021self,patel2020tailornet,bertiche2020cloth3d} rely on linear blend skinning w.r.t the underlying human skeleton to achieve garment deformations. 
While handling well in the tight-fitting garments such as shirts and trousers, they struggle with some loose-fitting garments (e.g., dresses and skirts) that deviate from body motions. 
Even though a few methods adopt dynamic skinning \cite{bertiche2021deepsd, bertiche2022neural, zhang2022motion} or virtual bones from skin decomposition \cite{pan2022predicting} , the predicted garments, especially skirt hem farther from the body, still appear stiff and unrealistic due to the limitation of linear representation.
Moreover, some methods \cite{wang2019learning, zhang2021dynamic} directly use neural networks to predict the deformations of loose-fitting garments, but their performance on complex motions and high-frequency wrinkles may not always be optimal.
In contrast, as shown in Figure \ref{fig1}, we present a data-driven method that simulates the free-flowing deformations of loose garments by explicitly learning the decoupled representation of garment deformation.

As analyzed above, the representation power of linear skinning can be limited, especially for loose-fitting garments that are highly nonlinear in nature.
Intuitively, deep neural networks may provide a suitable alternative due to their powerful nonlinear modelling capabilities.
However, the space of garment geometries is large, so directly using a network to fit the deformation from limited training samples often results in backfire.
To overcome this challenge, we build a garment generative model with the auto-encoder structure to learn a compact latent representation of garment geometries.
During decoding, we decompose the garment deformation using two separate decoders: one for reconstructing the unposed deformed garment and one for predicting per-vertex residual displacements (i.e., the offset difference between the garment in canonical pose and the posed garment), respectively.
Since the ground truth of canonical pose deformations is unavailable, we utilize sample points of the body surface to calculate the deformation representation and obtain the approximate unposed garment by the inverse transformation of human skinning for supervised training, which further introduces the pose prior guidance to help the decoder learn canonical deformations.
Note that the outputs of the two decoders are combined by addition to generate the posed garment geometry.
We then learn a mapping from the body motions and the previous states of the garment to this latent representation to estimate dynamic garment deformations.

To enhance the detailed deformations performance, we introduce a detail enhancement module to exploit local features of the generated garment and global latent representation features to estimate the high-frequency deformations.
We also adopt an adversarial setup by applying a multi-scale discriminator to obtain finer details while keeping global consistency.
For loose-fitting garments, we generate physical-based simulation data with varying motions of a digital avatar.
We evaluate the proposed method on different data and compare our results with state-of-the-art methods.
Experimental results demonstrate the superiority of our method both quantitatively and qualitatively.
We provide the code and video in the supplementary material.
In summary, our contribution are as follows:
\begin{itemize}
    \item We present a novel data-driven method to simulate large-scale deformations of loose-fitting garments without the limitation of linear skinning property and any specific body parameterization.
    \item We build a robust generative model with deformations decomposition to generate the posed garments from the latent space.
    We transfer the body motions and previous states to the
    latent space of out generative model to capture dynamic deformations.
    \item The proposed detail enhancement module with a multi-scale discriminator to improve high-frequency details.
\end{itemize}

%% file: sec/2_related.tex
\section{Related Work}
\label{sec:related}
In this section, we review previous work on garment animation
methods, which can be broadly divided into two categories: physics-based simulation and data-driven models.
\\
\textbf{Physics-Based Simulation.} Physics-based simulation methods have been widely studied in computer graphics, which tend to model realistic garment dynamics based on the material properties of garments.
The central studies include mechanical
modeling \cite{choi2005stable,nealen2006physically}, material modelling \cite{bhat2003estimating,miguel2012data}, using time integration \cite{baraff1998large,thomaszewski2008asynchronous} and collision handling \cite{provot1997collision,bridson2002robust}, etc.
While getting high-quality results, such PBS methods often have high computational costs. 
As a result, some methods \cite{muller2010wrinkle,rohmer2010animation} have been developed as constraint-based optimization methods to reduce computational efficiency.
\\
\textbf{Data-Driven Models.}
Recently, data-driven models based on deep learning have become popular alternatives to overcome the computationally expensive limitations of traditional physical simulators.
In contrast to PBS methods, data-driven methods, drawing on the observation of garment behavior in a dataset, learn a function that infers how garments deform under body motions.

One line of work predicts garment deformations using the deep neural network combined with linear blend skinning \cite{santesteban2019virtualtryon,patel2020tailornet,ma2020learning,bertiche2021pbns,gundogdu2019garnet}.
DeePSD \cite{bertiche2021deepsd} learns a mapping from the garment template mesh to its corresponding blend skinning weights.
SNUG \cite{santesteban2022snug} utilizes a GRU-based network to predict garment deformations in a canonical pose followed by skinning. 
Although these methods can produce realistic results, and some \cite{bertiche2021pbns, santesteban2022snug} even in an unsupervised way with reduced data demand, the skinning-based models limit the ability to handle loose, free-flowing garments such as dresses and skirts.

Some data-driven methods have also been explored to handle the deformation of loose-fitting garments.
A line of research \cite{bozic2021neural, tiwari2021neural, saito2021scanimate, ma2021scale, ma2021power} employs neural fields based implicit 
representations to model the clothed human. These methods, however, cannot model the garment separately, thus limiting the scope of use.
Instead of directly representing a clothed human, Santesteban et al. \cite{santesteban2021self} introduce a diffused body model by smoothly diffusing blend shapes and skinning weights to any 3D point around the body.
Bertiche et al. \cite{bertiche2022neural} disentangle static and dynamic garment deformations followed by dynamic skinning to model more free-flowing skirt hems.
Pan et al. \cite{pan2022predicting} create the virtual bones with per-vertex skinning weights via skin decomposition, to animate loose garments.
However, these methods still rely on linear skinning representation to achieve the deformation of loose garments w.r.t the underlying body, which limits their capability to cope with complex motions.


Another line of methods aims to learn a latent representation of plausible garment deformations from a dataset.
Wang et al. \cite{wang2019learning} learn an intrinsic garment space with a motion-driven auto-encoder network, which projects the garment mesh to latent space and decodes the garment using latent codes and the motion signature.
Su et al. \cite{su2022deepcloth} embed the UV-based representations into a continuous feature space, enabling plausible garment animation by leveraging the dynamic information encoded by shape and style representation.
Zhang et al. \cite{zhang2022motion} utilize a regularized auto-encoder to learn a generative space of garment deformations in the canonical pose.
However, these methods have limited representation capabilities for free-flowing deformations of loose-fitting garments.

Unlike previous methods, the proposed method does not directly depend on skinning and is able to model large-scale deformations of loose-fitting garments by learning a decoupled representation of garments geometries.

\begin{figure*}[!ht]
	\centering
	\includegraphics[width=\linewidth]{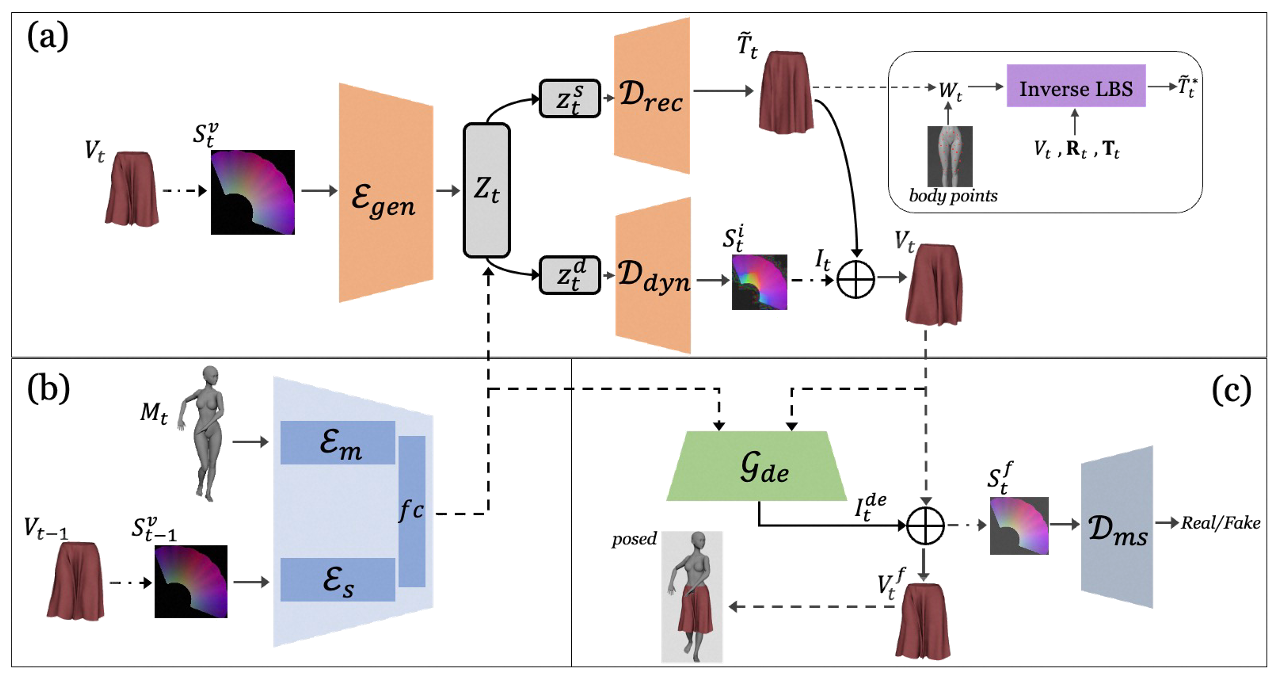}
	\caption{
        Overview of the proposed method. It mainly consists of three parts: (a) Garment Generative Model, (b)  Dynamic Motion Encoder and (c) Detail Enhancement Module.
	}
	\label{Fig2}            
\end{figure*}

%% file: sec/3_method.tex
\section{Method}

\subsection{Overview}
In this section, we present an overview of our method, as shown in Figure \ref{Fig2}.
Given a 3D character body $B$ and a specific garment template $T$, simulated under certain motions at the training stage, 
our method predicts how the garments would dynamically deform over the character's target (unseen) motion sequence $M_{t}$ (where $t$ is the frame number in the sequence).

Due to the highly nonlinear nature of garment dynamics, especially loose-fitting garments (e.g., skirts with long hems), it is challenging for the existing methods \cite{santesteban2021self,bertiche2022neural,santesteban2022snug,zhang2022motion} to simulate by directly predicting the deformations in canonical pose followed by the skinning.
Hence, we aim to first build a garment generative model that can effectively simulate highly nonlinear garment deformations. 
To this end, we learn a compact latent space of deformed garment geometries based on an auto-encoder network (Section \ref{section32}).
To simulate complex garment deformations, we decouple the latent code and decompose the deformation using two separate decoders. 
One decoder $\mathcal{D}_{rec}$ reconstructs the unposed deformed garment, and one decoder $\mathcal{D}_{dyn}$ predicts per-vertex residual displacements (i.e., the offset between the garment in canonical pose and the posed garment).
Since the ground truth of the unposed garments is unavailable, we exploit the skinning parameters of the body surface to project deformed garments into the canonical space to obtain approximate garment geometries.
Once we learn this generative model, in a subsequent stage, we train a dynamic motion encoder (Section \ref{section33}). 
We transfer body motions to garment generation by combining the motion encoder with the decoders of the generative model.
So far, our method can predict the deformed garment $V_{t}$ under the target motions. 
Next, we further learn high-frequency wrinkles based on the already generated garments with the help of the proposed detail enhancement module (Section \ref{section34}).
In the detail enhancement module, we introduce a patch-based multi-scale discriminator to obtain global consistency and finer details through adversarial training.




\subsection{Garment Generative Model}
\label{section32} 

Instead of predicting the local displacements in the canonical state of the garment and computing the posed deformation through linear skinning, we directly utilize the deep network to predict garment deformations.
We build an robust generative model based on an auto-encoder to represent the mapping from input garment $V_{t} \in \mathbb{R}^{n \times 3}$ to latent space.
Specifically, we project the 3D vertex positions of the input garment $V_{t}$ into the UV space to generate a 2D map $S_t^v \in \mathbb{R}^{w \times h \times 3}$, where $(w,h)$ are the dimensions of the UV map.
Encoding the position coordinates through the 2D UV map can exploit the locality of the 2D convolutions and help to capture the neighborhood information \cite{zhang2022motion}.
Given the input map $S_t^v$, we feed it into an encoder network $\mathcal{E}_{gen}$, composed of 2D convolutions layers and a fully-connected layer, to generate the latent code $Z_t \in \mathbb{R}^{128}$. The process is formulated as follows:
\begin{equation}
Z_t=\mathcal{E}_{gen}\left(S_t^v\right)
\end{equation}

Importantly, we expect the decoder to reconstruct garment $V_{t}$ from latent code $Z_t$, i.e. learn a mapping $Z_t \rightarrow V_{t}$.
A straightforward approach is to directly regress the vertices of the garment with a decoder symmetric to the encoder, but this suffers blending and smoothing artifacts, especially on loose garments (as we experiment in Section \ref{section42}). 
Therefore, we decompose the garment $V_{t}$ into unposed deformed mesh $\tilde{T_t}$ (in canonical pose) as well as the dynamic offset between $V_{t}$ and $\tilde{T_t}$:
\begin{equation}
V_t= \tilde{T_t} + I_t
\end{equation}
The offset $I_t$ is mainly due to the nonlinear displacement of the garment following the body movement (such as the floating skirt hem), while the unposed mesh $\tilde{T_t}$ represents the basic garment shape with some local displacements.
The two decomposed parts can be separated in the generative subspace of deformations.
Correspondingly, the latent code $Z_t$ is factorized into two disentangled variables: $z_t^s \in \mathbb{R}^{64}$ and $z_t^d \in \mathbb{R}^{64}$.
And we utilize two separate decoders, 
$\mathcal{D}_{rec}$ and $\mathcal{D}_{dyn}$, to reconstruct the unposed mesh $\tilde{T_t}$ and predict the offsets $I_t$, respectively.
By learning this decomposed representation, our model can accurately capture the dynamic nonlinear deformations of the loose garment.

\textbf{Decoder $\mathcal{D}_{rec}$.}
Given $z_t^s$, a multi fully-connected layers decoder $\mathcal{D}_{rec}$ is adopted to reconstruct the unposed deformed mesh $\tilde{T_t}$:
\begin{equation}
\tilde{T_t}=\mathcal{D}_{rec}\left(z_t^s\right)
\end{equation}

Since the ground truth of the unposed garment is unavailable, we seek to approximate the unposed mesh by the inverse skinning transformation of the input $V_{t}$.
Inspired by \cite{santesteban2021self,zhang2022motion}, we regularly sample $m$ seed points on the body surface associated with the garment to calculate rotation and translation $\left\{\mathbf{R}_t, \mathbf{T}_t\right\}$ with respect to the canonical body pose.
Then we apply a softmax function on the vertex of mesh $\tilde{T_t}$ and the body seed points in the canonical pose to compute per-frame linear skinning weights $W_t$.
Specifically, the skinning weight between a point $i$ of $\tilde{T_t}$ and a seed point $j$ of the body is calculated as:
\begin{equation}
w_t^{ij} = \frac{\exp \left(-d(i,j)\right) }
{\sum_{k=1}^{m} \exp \left(-d(i,k)\right)}
\end{equation}
where $d(\cdot)$ is the Euclidean distance function. 
Finally, we apply the inverse skinning with $\left\{V_{t}, \mathbf{R}_t, \mathbf{T}_t, W_t\right\}$ to obtain the unposed deformed mesh for the training, which further introduces the pose prior information to learn the local deformations of the unposed garment.



\textbf{Decoder $\mathcal{D}_{dyn}$. }
Given $z_t^d$, we use a decoder $\mathcal{D}_{dyn}$ with a
structure symmetrical to the encoder $\mathcal{E}_{gen}$ to generate the feature map $S_t^i$ with respect to residual offsets:
\begin{equation}
S_t^i=\mathcal{D}_{dyn}\left(z_t^d\right)
\end{equation}

Based on the inverse correspondence (from 3D to UV space projection), we smoothly sample from the feature map $S_t^i$ to obtain the vertex offset $I_{t}$. 
Finally, we add the offset $I_{t}$ to the generated deformed mesh $\tilde{T_t}$ to get the garment geometry $V_{t}$.
We present the detailed structure in Figure \ref{Fig2}(a).

\textbf{Training details.}
We train our generative model to learn the parameters of encoder and decoders in an end-to-end manner.

We apply a standard L2 loss between the generated mesh $V_{t}$ and the ground truth $V_t^*$ to make the generated $V_{t}$ close to the truth $V_t^*$.
\begin{equation}
L_{geo}=\left\|V_t-V_t^*\right\|_2+  \lambda_{0} \left\|\Delta(V_t)-\Delta (V_t^*)\right\|_2
\end{equation}
where $\Delta(\cdot)$ is the mesh Laplacian operator and $\lambda_{0}$ is a predefined weight factor.

For the decoder $\mathcal{D}_{rec}$, we first apply a loss term that enforces the reconstructed geometry $\tilde{T_t}$ to be similar to the approximate result $\tilde{T_t^*}$ obtained by inverse skinning.
We also utilize a small weighted loss term between $\tilde{T_t}$ and the garment template $T$, which enables the decoder $\mathcal{D}_{rec}$ to converge to the canonical pose state stably. We define them as:
\begin{equation}
L_{rec}= \lambda_{1}\left\|\tilde{T_t}-\tilde{T_t^*}\right\|_2+  \lambda_{2} \left\|\tilde{T_t}-T\right\|_2
\end{equation}
where $\lambda_{1}$ and $\lambda_{2}$ are weight factors.

To obtain a compact and continuous latent space in the generative model, we apply a KL regularization term \cite{vieillard2020leverage} to ensure the latent code consistent with the normal distribution $\mathcal{N}(0,1)$:
\begin{equation}
L_{reg}= \lambda_{3} KL\left(Z_t \| \mathcal{N}(0,1) \right)
\end{equation}
Finally, the total objective is a weighted sum of these loss terms given by: $L =L_{geo} + L_{rec} + L_{reg}$.

Once the training converges, we fix the weights of the encoder and decoders for the rest of the pipeline. 






\subsection{Dynamic Motion Encoder}
\label{section33}
In Section \ref{section32}, our garment generative model learns a mapping (i.e., $Z_t \rightarrow V_{t}$) from latent representation to plausible garment deformations, and we next predict the garment deformations at a frame $t$ based on the body motion $M_t$.
Considering the temporal coherence of garment deformations, we learn the garment dynamics from both the current motion $M_t$ and the previous states of the garment $V_{t-1}$.
Hence, we design a motion encoder that maps the body motion and the previous states of the garment to the latent space of the generative model, i.e. learn a mapping $\left\{M_t, V_{t-1} \right\} \rightarrow Z_t$.

Given the body motions $M_t$ (i.e., the rotations of joints and translations of the body), we feed it into the block $\mathcal{E}_m$, composed of a fully connected layer and a Gated Recurrent Unit (GRU) \cite{gers2000learning}, to obtain the motion vector.
For the previous garment frame $V_{t-1}$, we convert it into a 2D map $S_{t-1}^v$ as described in Section \ref{section32}, and adopt a block $\mathcal{E}_s$ similar to network $\mathcal{E}_{gen}$ to output the garment vector.
Then, we concatenate the two vectors and map them into the learned latent space through a fully-connected layer.
The whole process (see Figure \ref{Fig2}(b)) is as follows:
\begin{equation}
Z_t= fc\left(\mathcal{E}_m(M_t)\| \mathcal{E}_s(S_{t-1}^v)\right)
\end{equation}

After we obtain the latent code $Z_t$, we use the decoding process of the generative model to predict the posed garment geometry $V_t$.
During the training phase, we fix the weights of the decoders and train the motion encoder with the following loss function:
\begin{equation}
L_{geo}=\left\|V_t-V_t^*\right\|_2+  \lambda_{4} \left\|\Delta(V_t)-\Delta (V_t^*)\right\|_2
\end{equation}
\begin{equation}
L_{z}= \lambda_{5} \left\|Z_t-Z_t^*\right\|_2  
\end{equation}
\begin{equation}
\begin{aligned}
L &= L_{geo} + L_{z}
\end{aligned}
\end{equation}
where $V_t^*$ is the ground truth geometry for the current frame, and $Z_t^*$ is the corresponding latent code obtained by the generative model encoding the ground truth geometries.




\subsection{Detail Enhancement Module}
\label{section34}
Although our pipeline can effectively simulate plausible dynamic deformations of the garment, it is inevitable to smooth out some local high-frequency wrinkles in restoring garment geometry from latent code.
To this end, we introduce a detail enhancement module with an adversarial setup to estimate the detailed deformations of the garment.

As shown in Figure \ref{Fig2}(c), our network consists of a generator $\mathcal{G}_{de}$ and a discriminator $\mathcal{D}_{ms}$.
Taking the generated garment geometry $V_t$ and the latent code $Z_t$ as input, our generator generates the fine wrinkles by fusing local geometric deformations and global latent representation.
Specifically, we employ two graph convolution layers based on the EdgeConv operator \cite{wang2019dynamic} to extract local geometric features of the generated garment $V_t$. 
In parallel, we employ two full-connected layers to obtain global features from the latent code $Z_t$.
Then, we concatenate local and global features and feed the concatenated features into a fully-connected layer to obtain the high-frequency deformations $I_{t}^{de}$. 
The generated garment $V_t$ and deformations $I_{t}^{de}$ are combined into the final garment geometry $V_t^{f}$ by addition: $V_t^{f}= V_t + I_{t}^{de}$.

We design a patch-based multi-scale discriminator that consists of four 2D down-sampling convolutional layers.
Given the final generated garment $V_t^{f}$ and the ground truth $V_t^*$, we convert them into 2D maps i.e., $S_{t}^{f}$ and $S_{t}^{*}$ as described in Section \ref{section32}.
The discriminator takes $S_{t}^{f}$ or $S_{t}^{*}$ as input and identifies a patch of neighboring deformations as real or fake at multiple scales.
This enables the discriminator to inspect the deformations at different levels of granularity and guide the generator to obtain global consistency and fine local details.

During the training, we truncate the gradient of the generator input and train the generator and discriminator jointly.
We first consider an L1 loss w.r.t the output of the generator and the ground truth:
\begin{equation}
L_{geo}=\left\|V_t^{f}-V_t^*\right\|_1+ \left\|\Delta(V_t^{f})-\Delta (V_t^*)\right\|_1
\end{equation}

Furthermore, we adopt a least-squares \cite{mao2017least} adversarial loss to optimize $\mathcal{G}_{de}$ and $\mathcal{D}_{ms}$ jointly:
\begin{equation}
L_{g} = 
\mathbb{E}_{S^{*}}\left[(D_{ms}(S_{t}^{*})-1)^2\right]+ 
\mathbb{E}_{S^{f}} \left[(D_{ms}(S_{t}^{f})+1)^2\right]
\end{equation}
where the generator tries to minimize the loss against the discriminator to maximize it.
The overall loss function is the sum of these terms:
\begin{equation}
\begin{aligned}
L &= L_{geo}+ L_{g}
\end{aligned}
\end{equation}






\subsection{Collision Handling}
\label{section35}
During the inference, we observed that for unseen large motion sequences, some body-garment collisions occur.
We present an efficient optimization algorithm to push the garment outside the body while preserving its local shape.

Given the predicted garment $V_t^f$ under a body shape $B_t$, 
we first find the closest body point with position $b_t^k$ and normal $n_t^k$ for each vertex $v_t^k$ of the garment.
We then compute the projected distance between $v_t^k$ and $b_t^k$ on the vertex normal $n_t^k$ as: $ds = n_t^k \cdot \left(v_t^k-b_t^k\right)$.
Finally, we deform the garment mesh $V_t^f$ to obtain the collision-resolved garment $\tilde{V}_t^f$ by minimizing the following objective function:
\begin{equation}
L=\left\|\Delta(V_t^f)-  \Delta (\tilde{V}_t^f)\right\|_2 + 
\sum_{k \in V_t}max(\epsilon- ds, 0)
\end{equation}
where $\epsilon$ is a small positive threshold to prevent the garment from overlapping the body surface.



%% file: sec/4_expr.tex
\section{Experiments}
In this section, we evaluate our method qualitatively and quantitatively on different datasets to demonstrate the superiority of the method.
\subsection{Experimental Setup} 
\textbf{Data preparation. }
To test our method, we created a novel dataset for a skirt (9352 vertices) with a loose hem driven by a digital avatar. 
We select 15 motion sequences about 17K motion frames in total, derived from the mocap data of different large-scale dances, which have been manually modified by artists.
Next, we use the Style3D \footnote{https://www.style3d.com/}, a physically-based simulation system, to generate the ground truth mesh sequences.
And we use 13 clips (70\%) for training and 5 clips (30\%) for test.
Additionally, in order to verify our method on public datasets, we use the vto-dataset \cite{santesteban2019virtualtryon} based on the SMPL model \cite{loper2015smpl}.
For this dataset, we select the simulation results of a dress on a medium body shape ($\beta = 0$) and split 49 clips for training and 4 clips for test.
\\
\textbf{Evaluation metrics.}
To quantitatively evaluate the performance of our generated results, including garment deformations and dynamics, we use three metrics: Root Mean Squared Error (RMSE), Hausdorff distance as well as Spatio-Temporal Edge Difference (STED) \cite{vasa2010perception} measuring the difference in dynamics between generated and ground truth geometries.
\\
\textbf{Implementation details.}
In the garment generative model, the encoder $\mathcal{E}_{gen}$ composed of four 2D down-sampling convolutional layers and a fully-connected layer is used to map the input into a 128D latent code $Z_t$. 
We decompose into $z_t^s$ and $z_t^d$ using a fully-connected layer, respectively.
Then, the decoder $\mathcal{D}_{rec}$ comprises four fully connected layers, and the decoder $\mathcal{D}_{dyn}$ comprises a fully connected layer and four 2D up-sampling convolutional layers. 
In the dynamic motion encoder, we use a fully-connected layer 
 and a single GRU layer of size 500 for block $\mathcal{E}_m$, a similar structure as the encoder $\mathcal{E}_{gen}$ for block $\mathcal{E}_s$, and follow a fully-connected layer to obtain the latent space.
 For the generator $\mathcal{G}_{de}$ of the detail enhancement module, we use three stacked EdgeConv layers with vertex dimensions of $[3, 8, 10]$ to extract the local features and two fully-connected layers to obtain global features. 
 We concatenate these two features and feed them into a fully-connected layer, as mentioned in section \ref{section34}. 
 In adversarial training, we adopt three scales of patch sizes of 64$\times$64, 32$\times$32, and 16$\times$16 in the discriminator $\mathcal{D}_{ms}$.
We use instance normalization and leakyReLU for all convolutional layers and only leakyReLU for all linear layers.

We use PyTorch \cite{paszke2019pytorch} for implementation and train our method on an NVIDIA Tesla V100 GPU.
All networks are optimized using the Adam optimizer with a weight decay rate of 1e-3.
We first train the garment generative model for 500 epochs with a learning rate of 1e-4 and batch size of 56.
After that, we train the dynamic motion encoder for 300 epochs with a learning rate of 1e-4 and a batch size of 1.
To speed up the training process of GRU structure, we adopt Truncated Back propagation Through Time (TBPTT) with 50 time steps.
Finally, we train the detail enhancement network for 100 epochs with a learning rate of 2e-4 and batch size of 8.
The weight factors $\lambda_{0}$, $\lambda_{1}$, $\lambda_{2}$, $\lambda_{3}$, $\lambda_{4}$ and $\lambda_{5}$ are set to 3.5, 0.1, 0.05, 1e-3, 1.5, 0.5 respectively.
We set the threshold $\epsilon=$1e-3 in the collision handling.

\begin{figure}[t]
    \centering
    \includegraphics[width=0.9\linewidth]{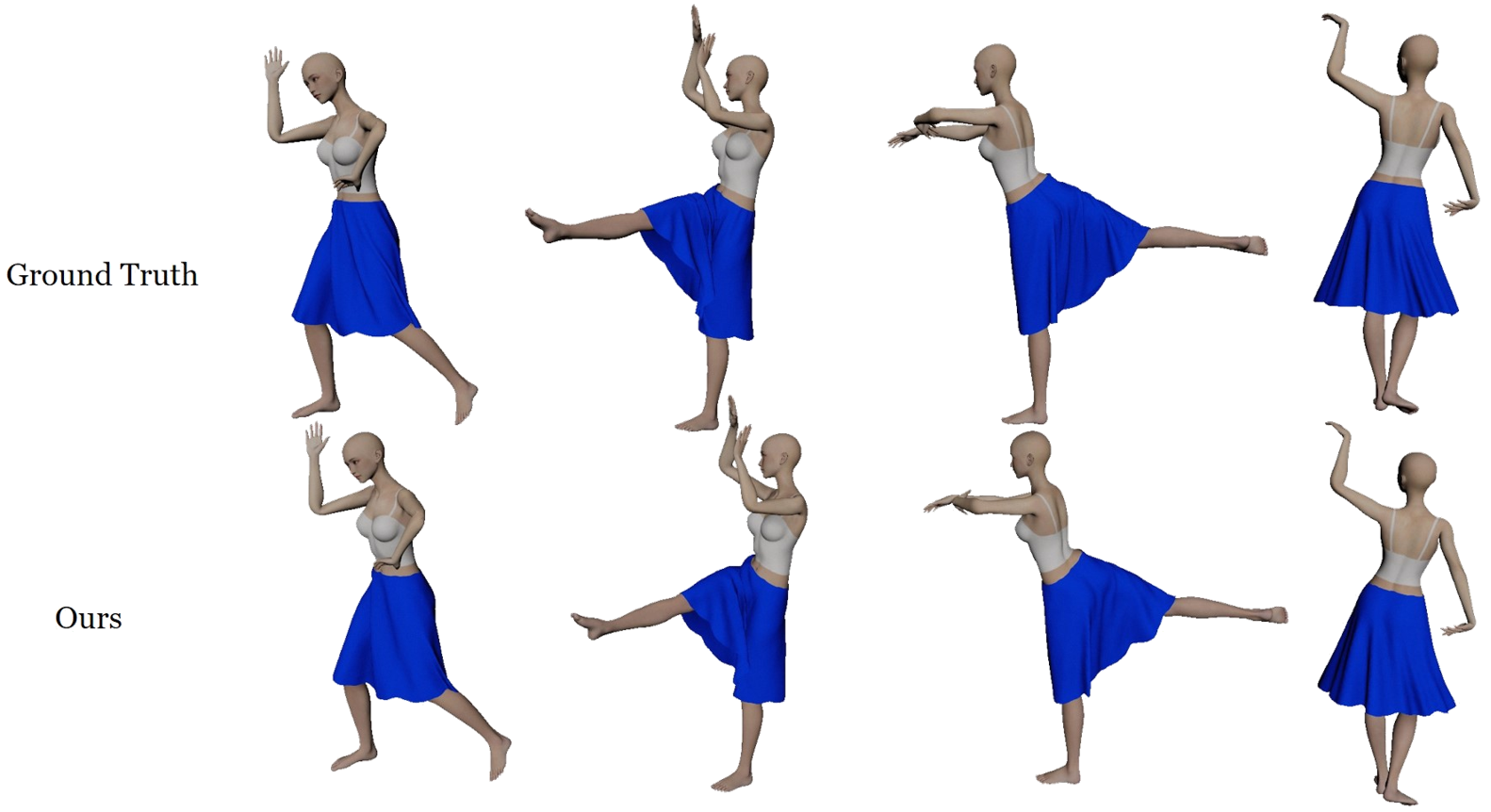}
    
   \caption{
   Given different dance motion sequences, our method can estimate high-quality garment animations close to the ground truth.
   }
\label{fig3}
\end{figure}

\begin{figure}[t]
    \centering
    \includegraphics[width=0.9\linewidth]{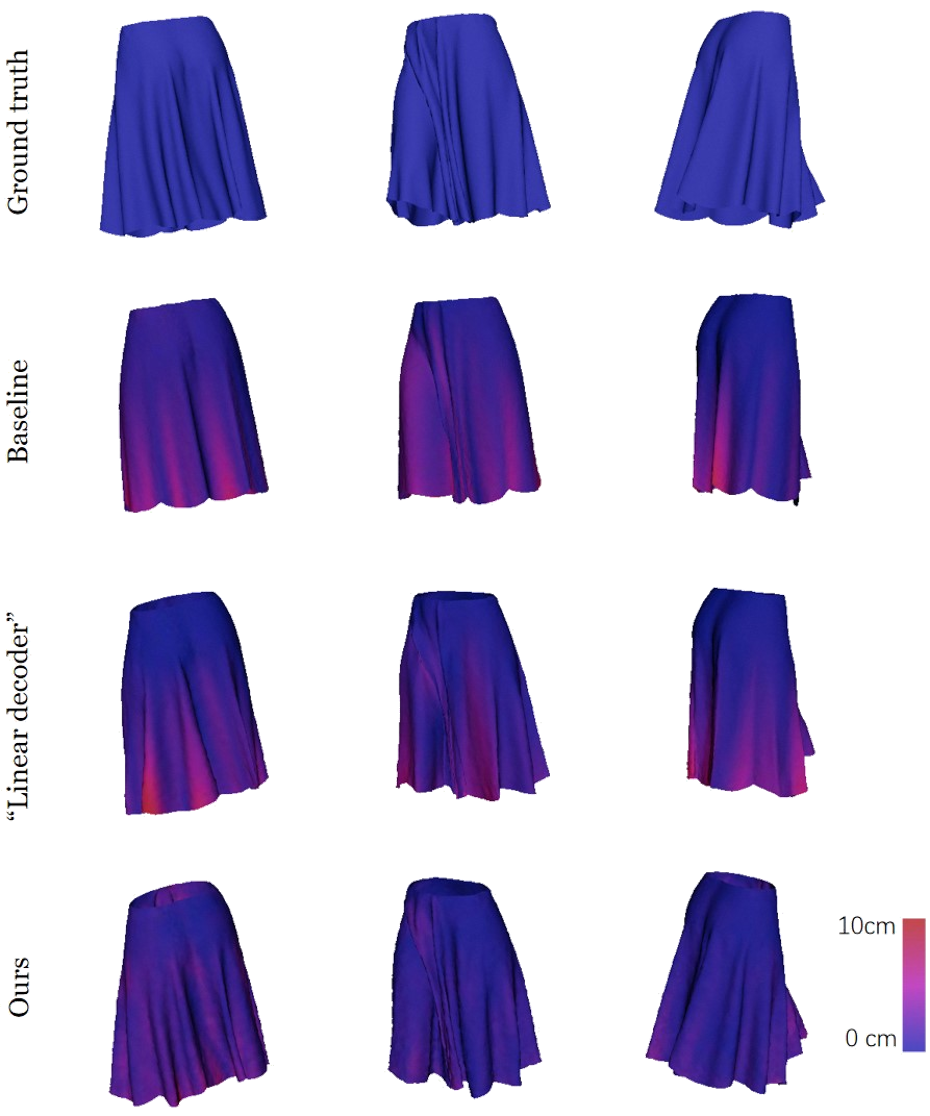}
    
   \caption{
   For a given garment under different poses (top row), we show the reconstruction results from three models: an auto-encoder with a single vertex decoder ($2^{nd}$ row), a linear decoder based on SSDR ($3^{rd}$ row), and our generative model ($4^{th}$ row).
   We render the per-vertex garment difference map. 
   Colors from blue to red indicate the increasing error.
   Our generative model provides better visual quality for shape reconstruction.
   }
\label{fig4}
\end{figure}

\begin{figure}[t]
    \centering
    \includegraphics[width=0.9\linewidth]{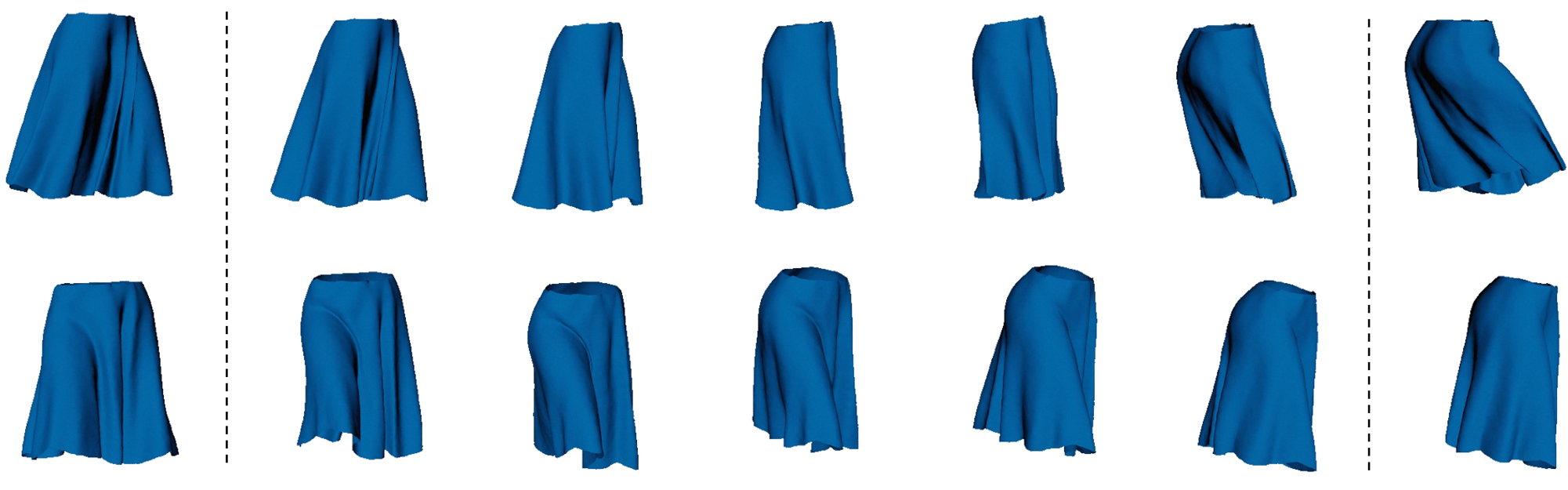}
    
   \caption{
   We perform uniform linear interpolation between the sampled two latent codes (leftmost and rightmost) and sequentially generate the corresponding garments shapes.
   We show different motion examples in two rows.
   The results indicate that our learned latent space is compact and continuous.
   }
\label{fig5}
\end{figure}

\subsection{Results and Evaluation}
\label{section42}
We first demonstrate the generalization capabilities of our method across unseen motion sequences.
As shown in Figure \ref{fig3}, we test several different types of dance sequences (including \emph{Ballet}, \emph{Jazz} and \emph{Folk dance}).
Our method generalizes well to motion sequences related to the distribution of poses seen during training.
For some exaggerated unseen motions (e.g., oversplit with legs), we observe unnatural garments deformations and heavy interpenetration with the body.
\\
\textbf{Garment Generative Model Evaluation.} 
As one key contribution, we learn a generative model based on deformation decomposition to model the mapping from latent space to dynamic garments.
We evaluate the nonlinear capacity of the generative model to garment deformations by measuring the reconstruction results.
As shown in Figure \ref{fig4}, we first adopt a baseline that directly predicts the absolute positions of garment vertices using a decoder with a symmetrical structure to the encoder $\mathcal{E}_{gen}$.
By designing two separate decoders to decouple unposed geometry and dynamic offsets, our model can effectively capture garment deformations (see the third row of Figure \ref{fig4}).
As an alternative method, we construct a ``linear decoder" that predicts deformations based on precomputed skinning decomposition (SSDR \cite{le2012smooth}) parameters.
However, we find that this method cannot fully capture the mesh, resulting in lower reconstruction accuracy compared with ours. 
Furthermore, to illustrate that the learned latent space is  compact and continuous in our generative model, we show garment deformations obtained by interpolating between two randomly sampled latent codes in Figure \ref{fig5}.
We see that the results produced in-between are plausible and change smoothly.
\\
\textbf{Motion Encoder Evaluation.} 
We train the encoder to learn $Z_t$ from $M_t, V_{t-1}$, and decode the current garment $V_t$. 
To verify the effect of predicting long sequences by using previous states of the garment iteratively, we evaluate the prediction results with and without previous states on multiple motion sequences.
As shown in the third and fifth rows of Table \ref{tab1}, our network can achieve lower error for sequences of more than a thousand frames, both in terms of geometric accuracy and temporal coherence.
And from short to long sequences, our results show better stability.
\\
\textbf{Detail Enhancement Evaluation.} 
We evaluate the role of the detail enhancement module on different motion sequences.
We compare the effect of applying the detail enhancement module and perform an ablation study that removes the discriminator with an adversarial setup.
The quantitative comparisons are shown in Table \ref{tab2}, where employing this module can enhance high-frequency wrinkles to improve the generated garments. 
Compared with no adversarial setup, our patch-based multi-scale discriminator guides the generator to obtain global consistency and finer details.





%


\begin{table}[ht]
\centering
\begin{threeparttable}
\begin{tabular}{c|c|ccc}
\toprule
Motion & Method & iter-100  & iter-500 & iter-1200 \\
\midrule
\multirow{2}{*}{Walk} 
&  w/o \emph{ps} & 0.89 & 0.91 & 0.95 \\
& with \emph{ps} & 0.85 & 0.86 & 0.88 \\
\midrule 
\multirow{2}{*}{Ballet} 
&  w/o \emph{ps} & 1.31 & 1.35 & 1.44 \\
& with \emph{ps} & 1.23 & 1.25 & 1.31 \\
\bottomrule
\end{tabular}
\end{threeparttable}
\caption{
To evaluate the performance of our motion encoder, we report the RMSE error with and without the previous state of garments on two different types of test sequences.
Note that we show iterative roll-out prediction from short sequences to long sequences of over a thousand frames. ``\emph{ps}" means the previous state, and ``iter-'' represents iteration frames.
}
\label{tab1}
\end{table}

\subsection{Comparison to State-of-the-Art Methods}
\begin{figure}[t]
    \centering
    \includegraphics[width=0.9\linewidth]{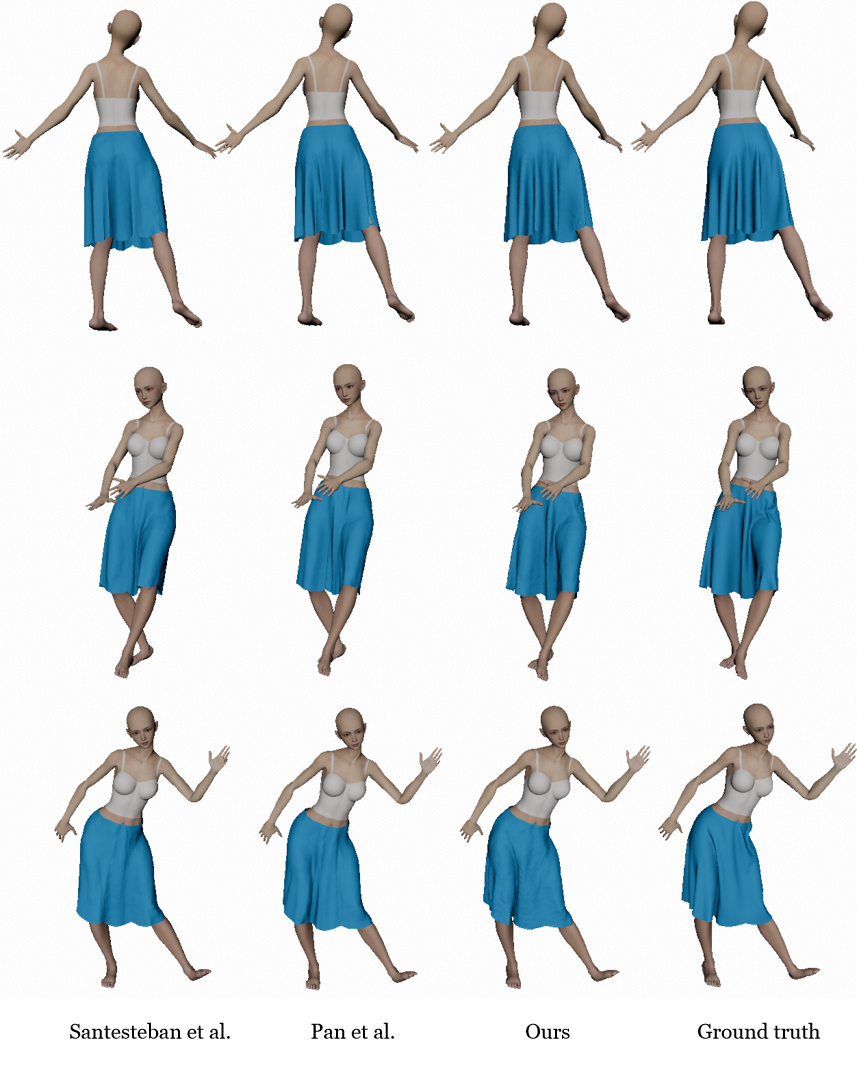}
    
   \caption{
    Qualitative comparison of prediction results with other methods. Our method generalizes to unseen motions to produce a more realistic performance.
   }
\label{fig6}
\end{figure}

To evaluate the performance of our method, we compare it to other state-of-the-art methods.
Specifically, we first perform a detailed comparison with Santesteban et al. \cite{santesteban2019virtualtryon} and Pan et al. \cite{pan2022predicting} (VB) on our skirt dataset.
Table \ref{tab3} presents the quantitative results on the test set.
Notice how our method consistently produces lower error across all three metrics, indicating that the results are more realistic and close to the ground truth.
We further present the visual quality comparisons.
As shown in Figure \ref{fig6}, the SOTA method VB is subject to skinning representation to result in a relatively stiff result for skirts with large-scale poses and cannot naturally keep the basic shape of the skirt on the dynamic motion.
In contrast, our method generates garment deformations that outperform it both in terms of realism and detail.

We also evaluate our method on public datasets containing tight garments. 
In Table \ref{tab4}, we provide quantitative results and show that our method improves over these methods.
We present a detailed example in Figure \ref{fig7}.
Our method estimates garment meshes with rich details and lower error.
Additional visual quality is much better demonstrated in the supplementary video.


\begin{table}[t]
\centering
\setlength{\belowcaptionskip}{-0.5cm}
\begin{threeparttable}
\begin{tabular}{l|ccc}
\toprule
Method & Walk & Ballet & Folk dance \\
\midrule
No module & 0.91 & 1.31 & 1.46 \\
module (w/o $\mathcal{D}_{ms}$) & 0.88 & 1.29 & 1.43 \\
Full module & 0.83 & 1.22 & 1.36 \\
\bottomrule
\end{tabular}
\end{threeparttable}
\caption{
To verify the effect of the detail enhancement module with an adversarial discriminator, we report the RMSE error of the prediction results on three different sequences.
}
\label{tab2}
\end{table}

\begin{table}[ht]
\centering
\setlength{\belowcaptionskip}{-0.5cm}
\begin{threeparttable}
\begin{tabular}{l|ccc}
\toprule
Method & RMSE $\downarrow$ & Hausdorff$ \downarrow$ & STED $\downarrow$ \\
\midrule
Santesteban et al. & 1.83 & 6.56 & 0.0872 \\
Pan et al. & 1.51 & 5.95 & 0.0789 \\
Ours & 1.13 & 5.17 & 0.0725 \\
\bottomrule
\end{tabular}
\end{threeparttable}
\caption{
Quantitative comparison of prediction garments for the unseen motions on the skirt dataset.
Our method produces lower errors on all metrics, indicating that our method results in better deformations and dynamics.
}
\label{tab3}
\end{table}

\begin{table}[ht]
\centering
\setlength{\belowcaptionskip}{-0.5cm}
\begin{threeparttable}
\begin{tabular}{l|ccc}
\toprule
Method & RMSE $\downarrow$ & Hausdorff$ \downarrow$ & STED $\downarrow$ \\
\midrule
Santesteban et al. & 2.096 & 8.70 & 0.0745 \\
Pan et al. & 1.991 & 8.34 & 0.0722 \\
Ours & 1.612 & 7.67 & 0.0687 \\
\bottomrule
\end{tabular}
\end{threeparttable}
\caption{
Compared with other methods for the overall performance on the vto-dataset set.
}
\label{tab4}
\end{table}

\begin{figure}[t]
\setlength{\belowcaptionskip}{-0.5cm} 
    \centering
    \includegraphics[width=0.8\linewidth]{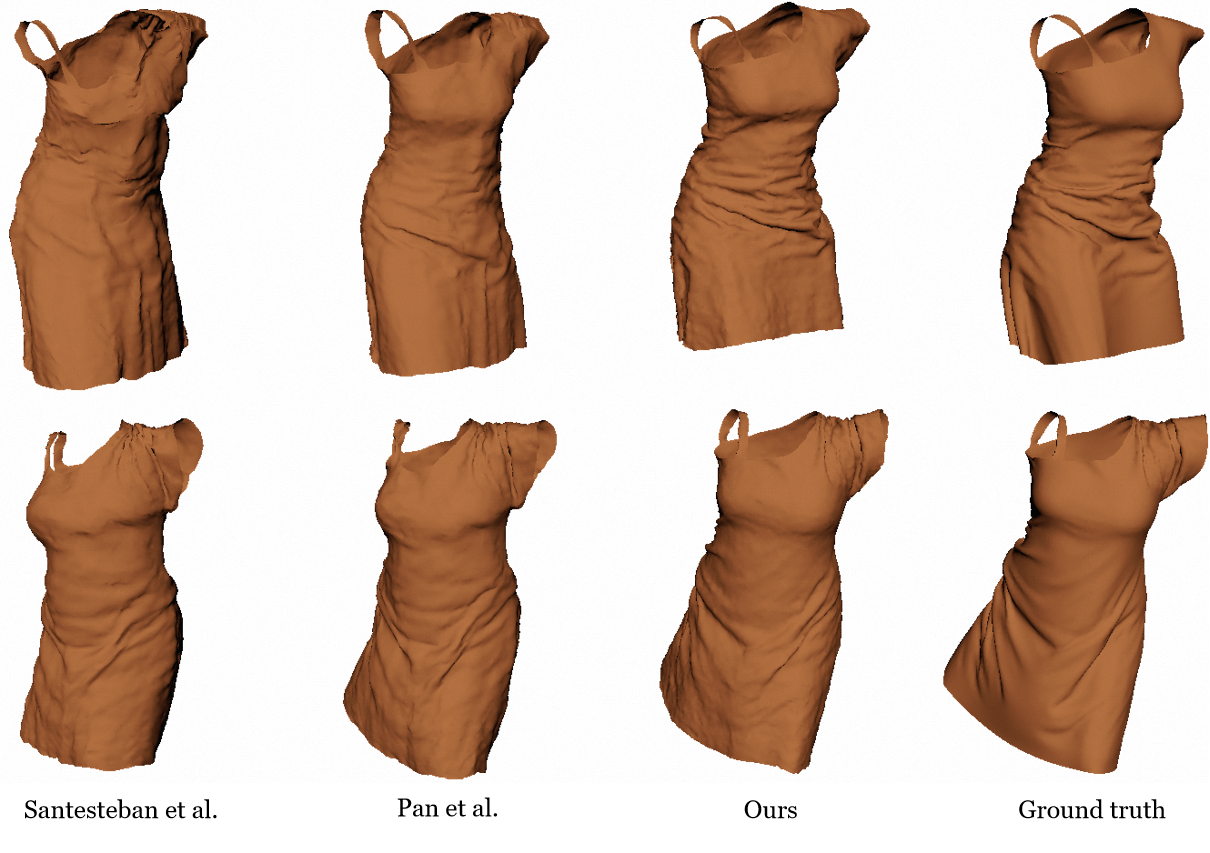}
    
   \caption{
    Qualitative comparison with other methods on the vto-dataset set.
   }
\label{fig7}
\end{figure}

%% file: sec/5_conc.tex
\section{Conclusion and Limitations}
In this paper, we present a novel data-driven method to learn complex dynamic deformations of loose-fitting garments.
Instead of relying on linear skinning, we propose a garment generative model, decomposing from the latent space of garment deformations into unposed geometry and dynamic offsets separately to simulate highly nonlinear garment deformations effectively.
We then introduce a dynamic motion encoder that transfers the body motions and the previous states of garment to latent space of our generative model to predict plausible dynamic deformations.
Finally, we improve the high-frequency details of the generated garment via a detail enhancement module with an adversarial setup.
We evaluate our method, both quantitatively and qualitatively,  and show improved performance compared to SOTA methods.

%

While showing realistic results, our method has some limitations.
Our method does not handle garment self-collisions caused by complex deformations, which needs to be addressed in future work.
Another limitation is that our current method is not able to 
deal with changes in topology, i.e. garment topology-free.
We observe that a graph-based network \cite{pfaff2020learning} may help solve this problem in the future.